\documentclass[10pt,twocolumn,letterpaper]{article}

\usepackage{iccv}
\usepackage{times}
\usepackage{epsfig}
\usepackage{graphicx}
\usepackage{amsmath}
\usepackage{amssymb}

\usepackage{multirow}
\usepackage{gensymb}
\usepackage{bm}
\usepackage{bbding}
\usepackage{subfigure}

\usepackage[pagebackref=true,breaklinks=true,letterpaper=true,colorlinks,bookmarks=false]{hyperref}

\iccvfinalcopy 

\def\iccvPaperID{4099} 

\ificcvfinal\fi

\begin{document}

\title{DualPoseNet: Category-level 6D Object Pose and Size Estimation \\ Using Dual Pose Network with Refined Learning of Pose Consistency}

\author{Jiehong Lin$^1$, Zewei Wei$^{1,2}$, Zhihao Li$^3$, Songcen Xu$^3$, Kui Jia$^1$\thanks{Corresponding author}, Yuanqing Li$^1$\\
$^1$South China University of Technology \\$^2$DexForce Technology Co. Ltd. $^3$Noah’s Ark Lab, Huawei Technologies Co. Ltd.\\
{\tt\small \{lin.jiehong, eeweizewei\}@mail.scut.edu.cn}, {\tt\small \{zhihao.li, xusongcen\}@huawei.com}, \\{\tt\small \{kuijia, auyqli\}@scut.edu.cn}
}


\maketitle
\ificcvfinal\thispagestyle{empty}\fi

\begin{abstract}
   Category-level 6D object pose and size estimation is to predict full pose configurations of rotation, translation, and size for object instances observed in single, arbitrary views of cluttered scenes. In this paper, we propose a new method of \emph{Dual Pose Network with refined learning of pose consistency} for this task, shortened as \emph{DualPoseNet}. DualPoseNet stacks two parallel pose decoders on top of a shared pose encoder, where the implicit decoder predicts object poses with a working mechanism different from that of the explicit one; they thus impose complementary supervision on the training of pose encoder. We construct the encoder based on spherical convolutions, and design a module of Spherical Fusion wherein for a better embedding of pose-sensitive features from the appearance and shape observations. Given no testing CAD models, it is the novel introduction of the implicit decoder that enables the refined pose prediction during testing, by enforcing the predicted pose consistency between the two decoders using a self-adaptive loss term. Thorough experiments on benchmarks of both category- and instance-level object pose datasets confirm efficacy of our designs. DualPoseNet outperforms existing methods with a large margin in the regime of high precision. Our code is released publicly at \url{https://github.com/Gorilla-Lab-SCUT/DualPoseNet}.
\end{abstract}

\section{Introduction}
\label{Sec: introduction}

Object detection in a 3D Euclidean space is demanded in many practical applications, such as augmented reality, robotic manipulation, and self-driving car. The field has been developing rapidly with the availability of benchmark datasets (e.g., KITTI \cite{KITTI} and SUN RGB-D \cite{SUNRGBD}), where carefully annotated 3D bounding boxes enclosing object instances of interest are prepared, which specify 7 degrees of freedom (7DoF) for the objects, including translation, size, and yaw angle around the gravity axis. This 7DoF setting of 3D object detection is aligned with common scenarios where the majority of object instances stand upright in the 3D space. However, 7DoF detection cannot precisely locate objects when the objects lean in the 3D space, where the most compact bounding boxes can only be determined given full pose configurations, \ie, with the additional two angles of rotation. Pose predictions of full configurations are important in safety-critical scenarios, e.g., autonomous driving, where the most precise and compact localization of objects enables better perception and decision making.

This task of pose prediction of full configurations (\ie, 6D pose and size) is formally introduced in \cite{NOCS} as \emph{category-level 6D object pose and size estimation} of novel instances from single, arbitrary views of RGB-D observations. It is closely related to \emph{category-level} amodal 3D object detection \cite{FPointNet,FConvNet, VoxelNet,PointRCNN,STD,VoteNet} (i.e., the above 7DoF setting) and \emph{instance-level} 6D object pose estimation \cite{PoseTemplateMatchingPAMI,drost2010model,hinterstoisser2016going, kehl2017ssd,xiang2017posecnn,sundermeyer2018implicit, peng2019pvnet, li2018unified, Densefusion, DeepIm}. Compared with them, the focused task in the present paper is more challenging due to learning and prediction in the full rotation space of $SO(3)$; more specifically, (1) the task is more involved in terms of both defining the category-level canonical poses (cf. Section \ref{SecProbStatement} for definition of canonical poses) and aligning object instances with large intra-category shape variations \cite{ShapeCollectionAlignment,ShapeNet}, (2) deep learning precise rotations arguably requires learning rotation-equivariant shape features, which is less studied compared with the 2D counterpart of learning translation-invariant image features, and (3) compared with instance-level 6D pose estimation, due to the lack of testing CAD models, the focused task cannot leverage the privileged 3D shapes to directly refine pose predictions, as done in \cite{ICP, xiang2017posecnn, Densefusion, DeepIm}.

\begin{figure*}[t]
    \begin{center}
       \vspace{-0.2cm}
       \includegraphics[width=0.98\linewidth]{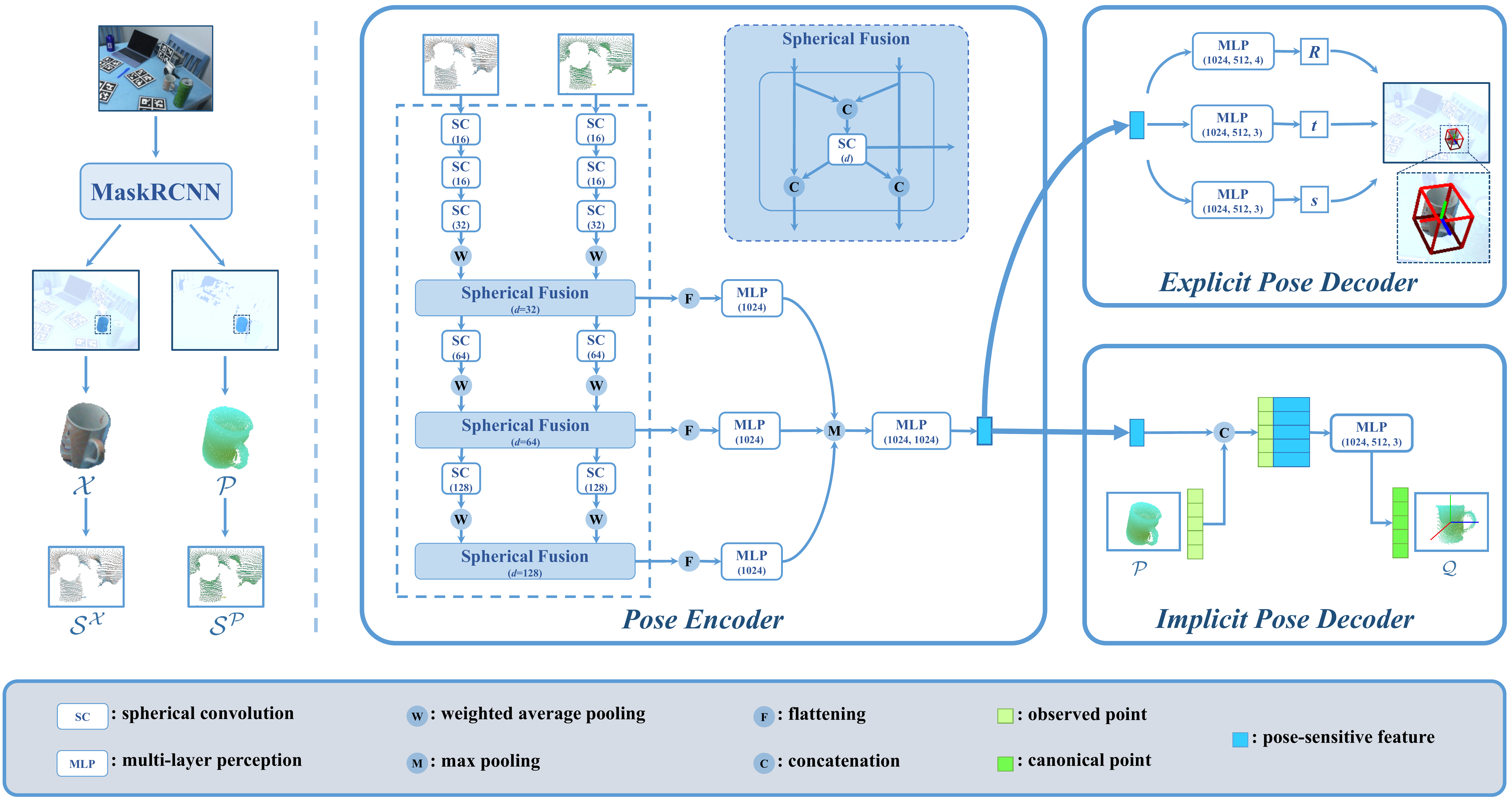}
    \end{center}
    \vspace{-0.3cm}
       \caption{{An illustration of our proposed DualPoseNet. For an observed RGB-D scene, DualPoseNet employs MaskRCNN \cite{ShapePrior} to segment out the object of interest, e.g., a mug, giving the observed points in $\mathcal{P}$ and the corresponding RGB values in $\mathcal{X}$, and feeds $(\mathcal{X}, \mathcal{P})$ into a \textbf{Pose Encoder} $\Phi$ to learn a pose-sensitive feature representation $\bm{f}$. Specifically, $\Phi$ is designed to have two parallel streams of spherical convolution layers to process the spherical signals $\mathcal{S}^{\mathcal{X}}$ and $\mathcal{S}^{\mathcal{P}}$ separately, which are respectively converted from $\mathcal{X}$ and $\mathcal{P}$; the resulting features are intertwined in the intermediate layers via a proposed module of \emph{Spherical Fusion}; finally $\bm{f}$ is enriched and obtained by aggregation of multi-scale spherical features.
       On top of $\Phi$, an \textbf{Explicit Pose Decoder} ${\Psi}_{exp}$ is constructed to directly predict the pose, while an additional \textbf{Implicit Pose Decoder} ${\Psi}_{im}$ is employed in parallel with ${\Psi}_{exp}$ to generate a canonical version $\mathcal{Q}$ of $\mathcal{P}$.
       }}
       \vspace{-0.35cm}
    \label{fig:netwrok}
\end{figure*}

In this work, we propose a novel method for category-level 6D object pose and size estimation, which can partially address the second and third challenges mentioned above. Our method constructs two parallel pose decoders on top of a shared pose encoder; the two decoders predict poses with different working mechanisms, and the encoder is designed to learn pose-sensitive shape features. A refined learning that enforces the predicted pose consistency between the two decoders is activated during testing to further improve the prediction. We term our method as \emph{Dual Pose Network with refined learning of pose consistency}, shortened as \emph{DualPoseNet}. Fig. \ref{fig:netwrok} gives an illustration.

For an observed RGB-D scene, DualPoseNet first employs an off-the-shelf model of instance segmentation (\eg, MaskRCNN \cite{MaskRCNN}) in images to segment out the objects of interest. It then feeds each masked RGB-D region into the encoder. To learn pose-sensitive shape features, we construct our encoder based on spherical convolutions \cite{SCNN,SCNNICLR}, which provably learn deep features of object surface shapes with the property of rotation equivariance on $SO(3)$. In this work, we design a novel module of \emph{Spherical Fusion} to support a better embedding from the appearance and shape features of the input RGB-D region. With the learned pose-sensitive features, the two parallel decoders either make a pose prediction \emph{explicitly}, or \emph{implicitly} do so by reconstructing the input (partial) point cloud in its canonical pose; while the first pose prediction can be directly used as the result of DualPoseNet, the result is further refined during testing by fine-tuning the encoder using a self-adaptive loss term that enforces the pose consistency. The use of implicit decoder in DualPoseNet has two benefits that potentially improve the pose prediction: (1) it provides an auxiliary supervision on the training of pose encoder, and (2) it is the key to enable the refinement given no testing CAD models. We conduct thorough experiments on the benchmark category-level object pose datasets of CAMERA25 and REAL275 \cite{NOCS}, and also apply our DualPoseNet to the instance-level ones of YCB-Video \cite{YCB} and LineMOD \cite{LineMod}. Ablation studies confirm the efficacy of our novel designs. DualPoseNet outperforms existing methods in terms of more precise pose. Our technical contributions are summarized as follows:
\begin{itemize}
    \item We propose a new method of \emph{Dual Pose Network} for category-level 6D object pose and size estimation. \emph{DualPoseNet} stacks two parallel pose decoders on top of a shared pose encoder, where the implicit one predicts poses with a working mechanism different from that of the explicit one; the two decoders thus impose complementary supervision on training of the pose encoder.

    \item In spite of the lack of testing CAD models, the use of implicit decoder in DualPoseNet enables a refined pose prediction during testing, by enforcing the predicted pose consistency between the two decoders using a self-adaptive loss term. This further improves the results of DualPoseNet.

    \item We construct the encoder of DualPoseNet based on spherical convolutions to learn pose-sensitive shape features, and design a module of \emph{Spherical Fusion} wherein, which is empirically shown to learn a better embedding from the appearance and shape features from the input RGB-D regions.
\end{itemize}

\section{Related Work}

\noindent \textbf{Instance-level 6D Object Pose Estimation} Traditional methods for instance-level 6D pose estimation include those based on template matching \cite{PoseTemplateMatchingPAMI}, and those by voting the matching results of point-pair features \cite{drost2010model,hinterstoisser2016going}. More recent solutions build on the power of deep networks and can directly estimate object poses from RGB images alone \cite{kehl2017ssd,xiang2017posecnn,sundermeyer2018implicit, peng2019pvnet} or RGB-D ones \cite{li2018unified, Densefusion}. This task assumes the availability of object CAD models during both the training and test phases, and thus enables a common practice to refine the predicted pose by matching the CAD model with the (RGB and/or point cloud) observations \cite{ICP,xiang2017posecnn,Densefusion,DeepIm}.

\noindent \textbf{Category-level 3D Object Detection} Methods for category-level 3D object detection are mainly compared on benchmarks such as KITTI \cite{KITTI} and SUN RGB-D \cite{SUNRGBD}. Earlier approach \cite{FPointNet,FConvNet} leverages the mature 2D detectors to first detect objects in RGB images, and learning of 3D detection is facilitated by focusing on point sets inside object frustums. Subsequent research proposes solutions \cite{VoxelNet,PointRCNN,STD,VoteNet} to predict the 7DoF object bounding boxes directly from the observed scene points. However, the 7DoF configurations impose inherent constrains on the precise rotation prediction, with only one yaw angle predicted around the gravity direction.

\noindent \textbf{Category-level 6D Object Pose and Size Estimation} More recently, category-level 6D pose and size estimation is formally introduced in \cite{NOCS}.
Notably, Wang \emph{et al.} \cite{NOCS} propose a canonical shape representation called normalized object coordinate space (NOCS), and inference is made by first predicting NOCS maps for objects detected in RGB images, and then aligning them with the observed object depths to produce results of 6D pose and size; later, Tian \emph{et al.} \cite{ShapePrior} improve the predictions of canonical object models by deforming categorical shape priors. Instead, Chen \emph{et al.} \cite{CASS} trained a variational auto-encoder (VAE) to capture pose-independent features, along with pose-dependent ones to directly predict the 6D poses. Besides, monocular methods are also explored in recent works \cite{chen2020category, manhardt2020cps++}.


\section{Problem Statement}
\label{SecProbStatement}
\vspace{-0.1cm}

Studies on category-level 6D object pose and size estimation start from NOCS \cite{NOCS}. The problem can be formally stated as follows. Assume a training set of cluttered scenes captured in RGB-D images, where ground-truth annotations of 6D pose and size for object instances of certain categories are provided. For each of the contained object instances, the annotation is in the form of full pose configuration of rotation $\bm{R} \in SO(3)$, translation $\bm{t} \in \mathbb{R}^3$, and size $\bm{s} \in \mathbb{R}^3$, which can also be translated as a compact, oriented 3D bounding box enclosing the object (cf. Fig. \ref{fig:netwrok}). Note that in a 3D Euclidean space, the 6D pose of $\bm{R}$ and $\bm{t}$ is defined relatively with respect to a \emph{canonical pose centered at the origin}. Category-level learning thus relies on the underlying assumption that all training object instances of a same category are aligned at a pre-defined canonical pose (\eg, handles of instances of a \textit{mug} category are all pointing towards a same direction); otherwise it makes no sense for any learned models to predict poses during testing. For existing datasets \cite{NOCS}, additional annotations of object masks in RGB images are usually provided, which eases the problem and enables learning to segment out regions of interest from RGB-D images of cluttered scenes.


\section{The Proposed Dual Pose Network with Refined Learning of Pose Consistency}
\label{SecMethod}

\subsection{Overview}
\label{SecOverview}

\begin{figure}[t]
    \begin{center}
    \vspace{-0.4cm}
       \includegraphics[width=0.9\linewidth]{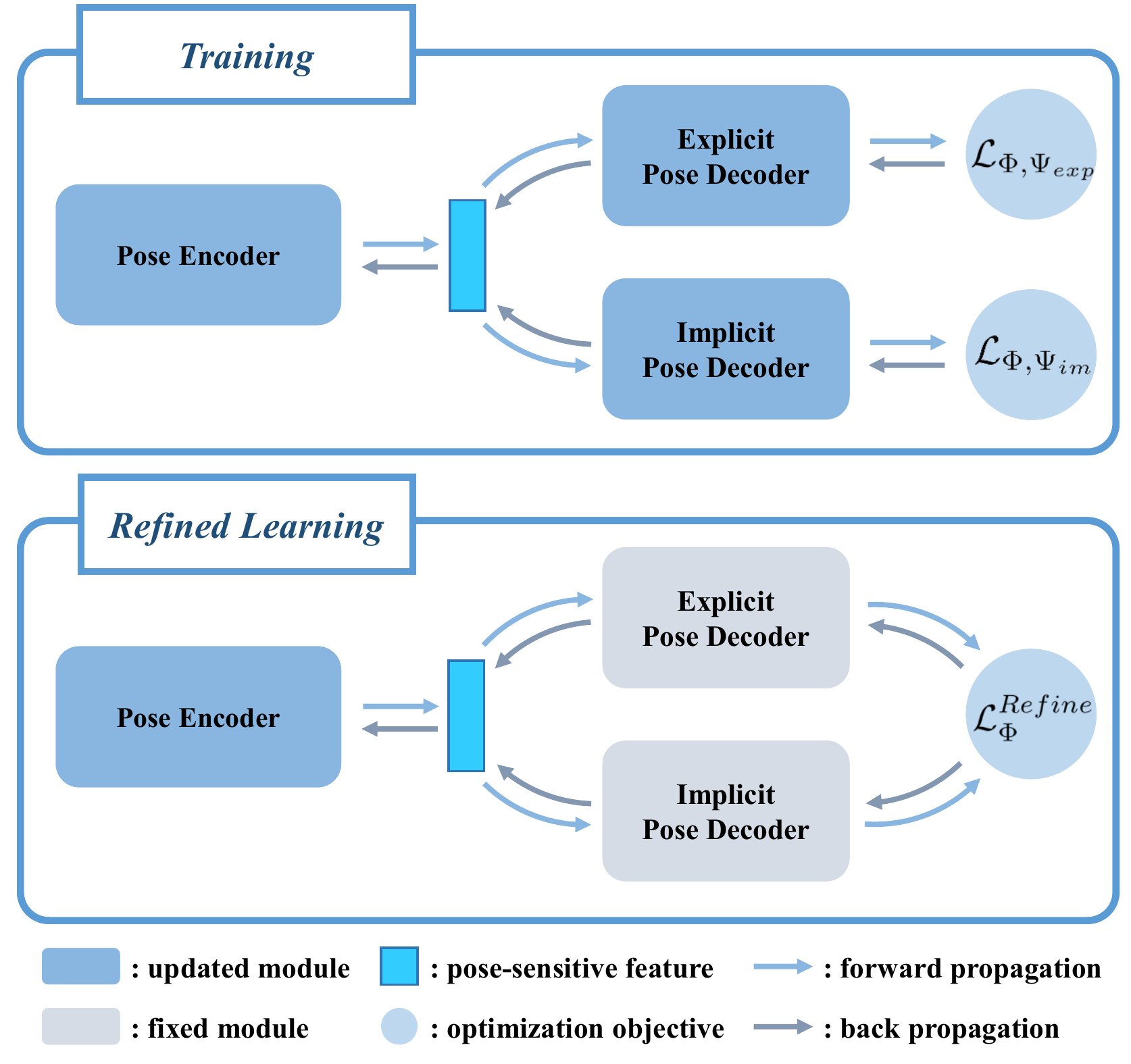}
    \end{center}
    \vspace{-0.2cm}
       \caption{{Illustrations on the training and refined learning of DualPoseNet. During training, we optimize the objective (\ref{EqnLossOverall}), which is a combination of $\mathcal{L}_{\Phi, \Psi_{exp}}$ and $\mathcal{L}_{\Phi, \Psi_{im}}$, in an end-to-end manner. During testing, we freeze the parameters of $\Psi_{exp}$ and $\Psi_{im}$, and fine-tune those of $\Phi$ to minimize $\mathcal{L}_{\Phi}^{Refine}$ for pose consistency. }}
    \vspace{-0.1cm}
    \label{fig:optimization}
\end{figure}

We first present an overview of our proposed \emph{Dual Pose Network with refined Learning of Pose Consistency}. The whole pipeline is depicted in Fig. \ref{fig:netwrok}. For an observed RGB-D scene, DualPoseNet first employs an off-the-shelf model of instance segmentation in images (\eg, MaskRCNN \cite{MaskRCNN}) to segment out the objects of interest. This produces a pair $(\mathcal{X}, \mathcal{P})$ for each segmented object, where we use $\mathcal{P} = \{\bm{p}_i \in \mathbb{R}^3 \}_{i=1}^{N}$ to represent the $N$ observed points in the masked RGB-D region and $\mathcal{X} = \{\bm{x}_i \in \mathbb{R}^3 \}_{i=1}^{N}$  for the corresponding RGB values. DualPoseNet feeds $(\mathcal{X}, \mathcal{P})$ into a \textbf{Pose Encoder} $\Phi$ (cf. Section \ref{Sec:9DoF Pose Encoder}) to learn a pose-sensitive feature representation $\bm{f}$, followed by an \textbf{Explicit Pose Decoder} ${\Psi}_{exp}$ (cf. Section \ref{Sec:Explicit Pose Decoder}) to predict the pose; an additional \textbf{Implicit Pose Decoder} ${\Psi}_{im}$ (cf. Section \ref{Sec:Implicit Pose Decoder}) is employed in parallel with ${\Psi}_{exp}$, which generates a canonical version $\mathcal{Q}$ of the observed point cloud $\mathcal{P}$. The use of ${\Psi}_{im}$ is the key in DualPoseNet to both improve the pose prediction from ${\Psi}_{exp}$, and enable a refined learning of pose consistency that further improves the precision of prediction, as verified in our experiments in Section \ref{Sec:Exp-Ablation}. Given cropped RGB-D regions of interest and the ground-truth pose annotations, training of DualPoseNet can be conducted in an end-to-end manner. During testing, there exist (at least) three ways to obtain the pose predictions from DualPoseNet: (1) the direct prediction from DualPoseNet via a forward pass of $\Psi_{exp} \circ \Phi$, (2) computing $\mathcal{Q} = \Psi_{im} \circ \Phi (\mathcal{X}, \mathcal{P})$ in its canonical pose and obtaining a pose prediction by solving Umeyama algorithm \cite{umeyama1991least} together with the observed $\mathcal{P}$, similar to existing methods \cite{NOCS,ShapePrior}, and (3) using the refined learning to update the parameters of the encoder $\Phi$, and then computing the prediction via a forward pass of $\Psi_{exp}\circ\Phi$. In this work, we use the first and third ways to obtain results of DualPoseNet. We illustrate the training and refinement processes in Fig. \ref{fig:optimization}. Individual components of the network are explained as follows.

\subsection{The Pose Encoder $\Phi$}
\label{Sec:9DoF Pose Encoder}

Precise prediction of object pose requires that the features learned by $\bm{f} = \Phi(\mathcal{X}, \mathcal{P})$ are sensitive to the observed pose of the input $\mathcal{P}$, especially to rotation, since translation and size are easier to infer from $\mathcal{P}$ (\eg, even simple localization of center point and calculation of 3D extensions give a good prediction of translation and scale). To this end, we implement our $\Phi$ based on spherical convolutions \cite{SCNN,SCNNICLR}, which provably learn deep features of object surface shapes with the property of rotation equivariance on $SO(3)$. More specifically, we design $\Phi$ to have two parallel streams of spherical convolution layers that process the inputs $\mathcal{X}$ and $\mathcal{P}$ separately; the resulting features are intertwined in the intermediate layers via a proposed module of \emph{Spherical Fusion}. We also use \emph{aggregation of multi-scale spherical features} to enrich the pose information in $\bm{f}$. Fig. \ref{fig:netwrok} gives the illustration.

\vspace{0.1cm}
\noindent \textbf{Conversion as Spherical Signals} Following \cite{SCNN}, we aim to convert $\mathcal{X}$ and $\mathcal{P}$ separately as discrete samplings $\mathcal{S}^\mathcal{X} \in \mathbb{R}^{W\times H\times 3}$ and $\mathcal{S}^\mathcal{P} \in \mathbb{R}^{W\times H\times 1}$ of spherical signals, where $W\times H$ represents the sampling resolution on the sphere. To do so, we first compute the geometric center $\bm{c} = \frac{1}{N} \sum_{i=1}^N \bm{p}_i$ of $\mathcal{P}$, and subtract its individual points from $\bm{c}$; this moves $\mathcal{P}$ into a space with the origin at $\bm{c}$. We then cast $W\times H$ equiangular rays from $\bm{c}$, which divide the space into $W \times H$ regions. Consider a region indexed by $(w, h)$, with $w \in \{1, \dots, W\}$ and $h \in \{1, \dots, H\}$; when it contains points of $\mathcal{P}$, we find the one with the largest distance to $\bm{c}$, denoted as $\bm{p}_{h,w}^{max}$, and define the spherical signal at the present region as $\mathcal{S}^\mathcal{X}(w, h) = \bm{x}_{h,w}^{max}$ and $\mathcal{S}^\mathcal{P}(w, h) = \| \bm{p}_{h,w}^{max} - \bm{c} \|$, where $\bm{x}_{h,w}^{max}$ denotes the RGB values corresponding to $\bm{p}_{h,w}^{max}$; otherwise, we define $\mathcal{S}^\mathcal{X}(w, h) = \bm{0}$ and $\mathcal{S}^\mathcal{P}(w, h) = 0$ when the region contains no points of $\mathcal{P}$.

\vspace{0.1cm}
\noindent \textbf{Learning with Spherical Fusion} As shown in Fig. \ref{fig:netwrok}, our encoder $\Phi$ is constructed based on two parallel streams that process the converted spherical signals $\mathcal{S}^\mathcal{X}$ and $\mathcal{S}^\mathcal{P}$ separately. We term them as $\mathcal{X}$-stream and $\mathcal{P}$-stream for ease of presentation. The two streams share a same network structure (except the channels of the first layers), each of which stacks multiple layers of spherical convolution and weighted average pooling, whose specifics are given in Fig. \ref{fig:netwrok}. To enable information communication and feature mixing between the two streams, we design a module of \emph{Spherical Fusion} that works as follows. Let $\mathcal{S}_l^\mathcal{X} \in \mathbb{R}^{W\times H\times d_l}$ and $\mathcal{S}_l^\mathcal{P} \in \mathbb{R}^{W\times H\times d_l}$ denote the learned spherical feature maps at the respective $l^{th}$ layers of the two streams (i.e., $\mathcal{S}_0^\mathcal{X} = \mathcal{S}^\mathcal{X}$ and $\mathcal{S}_0^\mathcal{P} = \mathcal{S}^\mathcal{P}$), we compute the input feature maps of layer $l+1$ for $\mathcal{P}$-stream as
\begin{eqnarray}
\widetilde{\mathcal{S}}_l^{\mathcal{P}} = \left[\mathcal{S}_l^{\mathcal{P}}, \widetilde{\mathcal{S}}^{\mathcal{X},\mathcal{P}}_l \right] \in \mathbb{R}^{W\times H\times 2d_l} \qquad\quad  \label{EqnSphericalFusion2} \\
\textrm{with} \ \ \widetilde{\mathcal{S}}^{\mathcal{X},\mathcal{P}}_l = \mathtt{SCONV}\left( \left[\mathcal{S}_l^{\mathcal{X}} , \mathcal{S}_l^{\mathcal{P}} \right] \right) \in \mathbb{R}^{W\times H\times d_l} , \label{EqnSphericalFusion1}
\end{eqnarray}
where $\texttt{SCONV}$ denotes a trainable layer of spherical convolution \cite{SCNN}, and $[\cdot, \cdot]$ concatenates spherical maps along the feature dimension. The same applies to $\mathcal{X}$-stream, and we have $\widetilde{\mathcal{S}}_l^{\mathcal{X}} \in \mathbb{R}^{W\times H\times 2d_l}$ as its input of layer $l+1$. The module of spherical fusion (\ref{EqnSphericalFusion2}) can be used in a plug-and-play manner at any intermediate layers of the two streams; we use three such modules between $\mathcal{X}$-stream and $\mathcal{P}$-stream of 5 spherical convolution layers for all experiments reported in this paper. Empirical analysis in Section \ref{Sec:Exp-Ablation} verifies the efficacy of the proposed spherical fusion. Note that a simple alternative exists that fuses the RGB and point features at the very beginning, i.e., a spherical signal $\mathcal{S} = [\mathcal{S}^\mathcal{X}, \mathcal{S}^\mathcal{P}] \in \mathbb{R}^{W\times H\times 4}$ converted from $(\mathcal{X}, \mathcal{P})$. Features in $\mathcal{S}$ would be directly fused in subsequent layers. Empirical results in Section \ref{Sec:Exp-Ablation} also verify that given the same numbers of spherical convolution layers and feature maps, this alternative is greatly outperformed by our proposed spherical fusion.

\vspace{0.1cm}
\noindent\textbf{Aggregation of Multi-scale Spherical Features} It is intuitive to enhance pose encoding by using spherical features at multiple scales. Since the representation $\widetilde{\mathcal{S}}_l^{\mathcal{X},\mathcal{P}}$ computed by (\ref{EqnSphericalFusion1}) fuses the appearance and geometry features at an intermediate layer $l$, we technically aggregate multiple of them from spherical fusion modules respectively inserted at lower, middle, and higher layers of the two parallel streams, as illustrated in Fig. \ref{fig:netwrok}. In practice, we aggregate three of such feature representations
as follows
\vspace{-0.1cm}
\begin{eqnarray}\label{EqnEncoderFea}
\bm{f} = \texttt{MLP}\left(\texttt{MaxPool}\left(\bm{f}_{l}, \bm{f}_{l'}, \bm{f}_{l''} \right)\right)  \qquad\qquad  \\
  \textrm{s.t.} \ \bm{f}_l = \texttt{MLP}\left(\texttt{Flatten}\left(\widetilde{\mathcal{S}}^{\mathcal{X},\mathcal{P}}_l\right)\right), \nonumber
\end{eqnarray}
where $\texttt{Flatten}(\cdot)$ denotes a flattening operation that reforms the feature tensor $\widetilde{\mathcal{S}}^{\mathcal{X},\mathcal{P}}_l$
of dimension $W\times H\times d_l$
as a feature vector, $\texttt{MLP}$ denotes a subnetwork of Multi-Layer Perceptron (MLP), and $\texttt{MaxPool}\left(\bm{f}_{l}, \bm{f}_{l'}, \bm{f}_{l''} \right)$ aggregates the three feature vectors by max-pooling over three entries for each feature channel;
layer specifics of the two MLPs used in (\ref{EqnEncoderFea}) are given in Fig. \ref{fig:netwrok}.
We use $\bm{f}$ computed from (\ref{EqnEncoderFea}) as the final output of the pose encoder $\Phi$, i.e., $\bm{f} = \Phi(\mathcal{X}, \mathcal{P})$.

\subsection{The Explicit Pose Decoder $\Psi_{exp}$}
\label{Sec:Explicit Pose Decoder}

Given $\bm{f}$ from the encoder $\Phi$, we implement the explicit decoder $\Psi_{exp}$ simply as three parallel MLPs that are trained to directly regress the rotation $\bm{R}$, translation $\bm{t}$, and size $\bm{s}$. Fig. \ref{fig:netwrok} gives the illustration, where layer specifics of the three MLPs are also given.
This gives a direct way of pose prediction from a cropped RGB-D region as $(\bm{R}, \bm{t}, \bm{s}) = \Psi_{exp} \circ \Phi(\mathcal{X}, \mathcal{P})$.

\subsection{The Implicit Pose Decoder $\Psi_{im}$}
\label{Sec:Implicit Pose Decoder}

For the observed point cloud $\mathcal{P}$, assume that its counterpart $\mathcal{Q}$ in the canonical pose is available. An affine transformation $(\bm{R}, \bm{t}, \bm{s})$ between $\mathcal{P}$ and $\mathcal{Q}$ can be established, which computes $\bm{q} = \frac{1}{||\bm{s}||}\bm{R}^T(\bm{p} - \bm{t})$ for any corresponding pair of $\bm{p} \in \mathcal{P}$ and $\bm{q} \in \mathcal{Q}$. This implies an implicit way of obtaining predicted pose by learning to predict a canonical $\mathcal{Q}$ from the observed $\mathcal{P}$; upon prediction of $\mathcal{Q}$, the pose $(\bm{R}, \bm{t}, \bm{s})$ can be obtained by solving the alignment problem via Umeyama algorithm \cite{umeyama1991least}. Since $\bm{f} = \Phi(\mathcal{X}, \mathcal{P})$ has learned the pose-sensitive features, we expect the corresponding $\bm{q}$ can be estimated from $\bm{p}$ by learning a mapping from the concatenation of $\bm{f}$ and $\bm{p}$. In DualPoseNet, we simply implement the learnable mapping as
\vspace{-0.2cm}
\begin{equation}\label{EqnImplicitDecoder}
\Psi_{im}(\bm{p}, \bm{f}) = \texttt{MLP}\left(\left[\bm{p}; \bm{f}\right]\right) .
\vspace{-0.2cm}
\end{equation}
$\Psi_{im}$ applies to individual points of $\mathcal{P}$ in a point-wise manner. We write collectively as $\mathcal{Q} = \Psi_{im}(\mathcal{P}, \bm{f})$.

We note that an equivalent representation of normalized object coordinate space (NOCS) is learned in \cite{NOCS} for a subsequent computation of pose prediction. Different from NOCS, we use $\Psi_{im}$ in an implicit way; it has two benefits that potentially improve the pose prediction: (1) it provides an auxiliary supervision on the training of pose encoder $\Psi$ (note that the training ground truth of $\mathcal{Q}$ can be transformed from $\mathcal{P}$ using the annotated pose and size), and (2) it enables a refined pose prediction by enforcing the consistency between the outputs of $\Psi_{exp}$ and $\Psi_{im}$, as explained shortly in Section \ref{SecDualPoseNetRefine}. We empirically verify both the benefits in Section \ref{Sec:Exp-Ablation}, and show that the use of $\Psi_{im}$ improves pose predictions of $\Psi_{exp} \circ \Phi(\mathcal{X}, \mathcal{P})$ in DualPoseNet.

\subsection{Training of Dual Pose Network}
\label{SecDualPoseNetTrain}

Given the ground-truth pose annotation $(\bm{R}^{*}, \bm{t}^{*}, \bm{s}^{*})$ \footnote{Following \cite{ShapePrior}, we use canonical $\bm{R}^{*}$ for symmetic objects to handle ambiguities of symmetry.} for a cropped $(\mathcal{X}, \mathcal{P})$, we use the following training objective on top of the explicit decoder $\Psi_{exp}$:
\begin{equation}\label{EqnLossExp}
    \mathcal{L}_{\Phi, \Psi_{exp}} =  ||\rho(\bm{R}) - \rho(\bm{R}^{*})||_2 + ||\bm{t} - \bm{t}^{*}||_2 + ||\bm{s} - \bm{s}^{*}||_2 ,
\end{equation}
where $\rho(\bm{R})$ is the quaternion representation of rotation $\bm{R}$.

Since individual points in the predicted $\mathcal{Q} = \{ \bm{q}_i \}_{i=1}^N$ from $\Psi_{im}$ are respectively corresponded to those in the observed $\mathcal{P} = \{ \bm{p}_i \}_{i=1}^N$, we simply use the following loss on top of the implicit decoder
\begin{equation}\label{EqnLossIm}
\mathcal{L}_{\Phi, \Psi_{im}} = \frac{1}{N} \sum_{i=1}^{N} \left\| \bm{q}_i - \frac{1}{|| \bm{s}^{*}||} \bm{R}^{*\top}(\bm{p}_i- \bm{t}^{*}) \right\|_2 .
\end{equation}

The overall training objective combines (\ref{EqnLossExp}) and (\ref{EqnLossIm}), resulting in the optimization problem
\begin{equation}\label{EqnLossOverall}
\min_{\Phi, \Psi_{exp}, \Psi_{im}} \mathcal{L}_{\Phi, \Psi_{exp}} + \lambda \mathcal{L}_{\Phi, \Psi_{im}} ,
\end{equation}
where $\lambda$ is a penalty parameter.

\subsection{The Refined Learning of Pose Consistency}
\label{SecDualPoseNetRefine}

For instance-level 6D pose estimation, it is a common practice to refine an initial or predicted pose by a post-registration \cite{ICP} or post-optimization \cite{DeepIm}; such a practice is possible since CAD model of the instance is available, which can guide the refinement by matching the CAD model with the (RGB and/or point cloud) observations. For our focused category-level problem, however, CAD models of testing instances are not provided. This creates a challenge in case that more precise predictions on certain testing instances are demanded.

Thanks to the dual pose predictions from $\Psi_{exp}$ and $\Psi_{im}$, we are able to make a pose refinement by learning to enforce their pose consistency. More specifically, we freeze the parameters of $\Psi_{exp}$ and $\Psi_{im}$, while fine-tuning those of the encoder $\Phi$, by optimizing the following problem
\vspace{-0.1cm}
\begin{equation}\label{EqnLossRefine}
\min_{\Phi} \mathcal{L}_{\Phi}^{Refine} = \frac{1}{N} \sum_{i=1}^{N} \left\| \bm{q}_i - \frac{1}{|| \bm{s}||} \bm{R}^{\top}(\bm{p}_i- \bm{t}) \right\|_2 ,
\end{equation}
where $\mathcal{Q} = \{\bm{q}_i\}_{i=1}^N = \Psi_{im}\circ \Phi(\mathcal{X}, \mathcal{P})$ and $(\bm{R}, \bm{t}, \bm{s})  = \Psi_{exp}\circ \Phi(\mathcal{X}, \mathcal{P})$ are the outputs of the two decoders. Note that during training, the two decoders are consistent in terms of pose prediction, since both of them are trained to match their outputs with the ground truths. During testing, due to an inevitable generalization gap, inconsistency between outputs of the two decoders always exists, and our proposed refinement (\ref{EqnLossRefine}) is expected to close the gap. An improved prediction relies on a better pose-sensitive encoding $\bm{f} = \Phi(\mathcal{X}, \mathcal{P})$; the refinement (\ref{EqnLossRefine}) thus updates parameters of $\Phi$ to achieve the goal. Empirical results in Section \ref{Sec:Exp-Ablation} verify that the refined poses are indeed towards more precise ones.
In practice, we set a loss tolerance $\epsilon$ as the stopping criterion when fine-tuning $\mathcal{L}_{\Phi}^{Refine}$ (\ie, the refinement stops when $\mathcal{L}_{\Phi}^{Refine} \leq \epsilon$), with fast convergence and negligible cost.


\section{Experiments}
\label{SecExps}

\noindent \textbf{Datasets} We conduct experiments using the benchmark CAMERA25 and REAL275 datasets \cite{NOCS} for category-level 6D object pose and size estimation. CAMERA25 is a synthetic dataset generated by a context-aware mixed reality approach from $6$ object categories; it includes $300,000$ composite images of $1,085$ object instances, among which $25,000$ images of $184$ instances are used for evaluation. REAL275 is a more challenging real-world dataset captured with clutter, occlusion and various lighting conditions; its training set contains $4,300$ images of $7$ scenes, and the test set contains $2,750$ images of $6$ scenes. Note that CAMERA25 and REAL275 share the same object categories, which enables a combined use of the two datasets for model training, as done in \cite{NOCS, ShapePrior}.

We also evaluate the advantages of DualPoseNet on the benchmark instance-level object pose datasets of YCB-Video \cite{YCB} and LineMOD \cite{LineMod}, which consist of $21$ and $13$ different object instances respectively.

\vspace{0.1cm}
\noindent \textbf{Implementation Details} We employ a MaskRCNN \cite{MaskRCNN} implemented by \cite{matterport_maskrcnn_2017} to segment out the objects of interest from input scenes. For each segmented object, its RGB-D crop is converted as spherical signals with a sampling resolution $64 \times 64$, and is then fed into our DualPoseNet. Configurations of DualPoseNet, including channel numbers of spherical convolutions and MLPs, have been specified in Fig. \ref{fig:netwrok}. We use ADAM to train DualPoseNet, with an initial learning rate of $0.0001$. The learning rate is halved every $50,000$ iterations until a total number of $300,000$ ones. We set the batch size as $64$, and the penalty parameter in Eq. (\ref{EqnLossOverall}) as  $\lambda = 10$.
For refined learning of pose consistency, we use a learning rate $1\times 10^{-6}$ and a loss tolerance $\epsilon=5\times10^{-5}$. For the instance-level task, we additionally adopt a similar 2nd-stage iterative refinement of residual pose as \cite{Densefusion, xu2019w} did; more details are shown in the supplementary material.

\vspace{0.1cm}
\noindent \textbf{Evaluation Metrics} For category-level pose estimation, we follow \cite{NOCS} to report mean Average Precision (mAP) at different thresholds of intersection over union (IoU) for object detection, and mAP at $n\degree$~$m$ cm for pose estimation. However, those metrics are not precise enough to simultaneously evaluate 6D pose and object size estimation, since IoU alone may fail to characterize precise object poses (a rotated bounding box may give a similar IoU value). To evaluate the problem nature of simultaneous predictions of pose and size, in this work, we also propose a new and more strict metric based on a combination of \emph{IoU}, \emph{error of rotation}, and \emph{error of relative translation}, where for the last one, we use the relative version since absolute translations make less sense for objects of varying sizes. For the three errors, we consider respective thresholds of $\{50\%, 75\%\}$ (\ie, IoU$_{50}$ and IoU$_{75}$), $\{5\degree, 10\degree\}$, and $\{5\%, 10\%, 20\%\}$, whose combinations can evaluate the predictions across a range of precisions. For instance-level pose estimation, we follow \cite{Densefusion} and evaluate the results of YCB-Video and LineMOD datasets by ADD-S and ADD(S) metrics, respectively.

\subsection{Category-level 6D Pose and Size Estimation}
\subsubsection{Ablation Studies and Analyses} \label{Sec:Exp-Ablation}

\begin{table*}
   \begin{center}
       \vspace{-0.1cm}
       \resizebox{1.0\textwidth}{!}{
       \begin{tabular}{c|c|c|cccccc|cccccc}
           \hline
           \multirow{3}{*}{Encoder} & \multirow{3}{*}{$\Psi_{im}$} & \multirow{3}{*}{Refining} & \multicolumn{12}{c}{mAP} \\
           \cline{4-15}
           & & & IoU$_{75}$ & IoU$_{75}$ & IoU$_{75}$ & IoU$_{50}$ & IoU$_{50}$ & IoU$_{50}$ & \multirow{2}{*}{IoU$_{50}$} & \multirow{2}{*}{IoU$_{75}$} & 5\degree & 5\degree & 10\degree & 10\degree\\
           & & & 5\degree, $5\%$ & 10\degree, $5\%$ & 5\degree, $10\%$ & 5\degree, $20\%$ & 10\degree, $10\%$ & 10\degree, $20\%$ & & & 2cm & 5cm & 2cm & 5cm \\
           \hline
           \hline
           Densefusion \cite{Densefusion} & \checkmark  & $\times$ & $1.5$ & $3.0$ & $7.9$ & $11.4$ & $17.4$ & $26.1$ & $64.9$ & $35.0$ & $9.1$ & $15.6$ &$19.3$ & $36.2$ \\
           SCNN-EarlyFusion & \checkmark & $\times$ & $7.7$ & $14.4$ & $15.8$ & $20.3$ & $35.5$ & $45.8$ & $76.1$ & $51.9$ & $17.3$ & $24.5$ & $36.2$ & $56.8$ \\
           SCNN-LateFusion & \checkmark & $\times$  & $8.4$ & $14.7$ & $23.8$ & $28.5$ & $41.7$ & $51.4$ & $77.0$ & $56.6$ & $25.7$ & $34.3$ & $43.5$ & $62.8$ \\
           \hline
           \hline
           $\Phi$ & $\times$ & $\times$ & $8.2$          & $13.6$          & $19.7$          & $26.1$          & $37.3$          & $49.1$ & $76.1$ & $55.2$ & $21.3$ & $31.3$ & $38.5$ & $60.4$\\
           $\Phi$ &  \checkmark   &   $\times$     & $10.4$& $16.1$ & $23.8$ & $28.5$ & $43.1$ & $52.6$ & $79.7$         & $60.1$          & $28.0$          & $34.3$          & $47.8$          & $64.2$\\
           $\Phi$ & \checkmark & \checkmark            & $\mathbf{11.2}$& $\mathbf{17.2}$ & $\mathbf{24.8}$ & $\mathbf{29.8}$ & $\mathbf{44.5}$     & $\mathbf{55.0}$ & $\mathbf{79.8}$& $\mathbf{62.2}$ & $\mathbf{29.3}$ & $\mathbf{35.9}$ & $\mathbf{50.0}$ & $\mathbf{66.8}$\\
           \hline
       \end{tabular}
       }
   \end{center}
   \vspace{-0.2cm}
   \caption{Ablation studies on variants of our proposed DualPoseNet on REAL275. Evaluations are based on both our proposed metrics (left) and the metrics (right) proposed in \cite{NOCS}.}\label{Table:Ablation}
   \vspace{-0.2cm}
\end{table*}

We first conduct ablation studies to evaluate the efficacy of individual components proposed in DualPoseNet. These studies are conducted on the REAL275 dataset \cite{NOCS}. 

We use both $\Psi_{exp}$ and $\Psi_{im}$ for pose decoding from DualPoseNet; $\Psi_{exp}$ produces the pose predictions directly, which are also used as the results of DualPoseNet both with and without the refined learning, while $\Psi_{im}$ is an implicit one whose outputs can translate as the results by solving an alignment problem. To verify the usefulness of $\Psi_{im}$, we report the results of DualPoseNet with or without the use of $\Psi_{im}$ in Table \ref{Table:Ablation}, in terms of the pose precision from $\Psi_{exp}$ \emph{before the refined learning}. We observe that the use of $\Psi_{im}$ improves the performance of $\Psi_{exp}$ by large margins under all the metrics; for example, the mAP improvement of (IoU$_{50}, 10\degree, 10\%$) reaches $5.8\%$, and that of (IoU$_{75}, 5\degree, 10\%$) reaches $4.1\%$. These performance gains suggest that $\Psi_{im}$ not only enables the subsequent refined learning of pose consistency, but also provides an auxiliary supervision on the training of pose encoder $\Phi$ and results in a better pose-sensitive embedding, implying the key role of $\Psi_{im}$ in DualPoseNet.

To evaluate the efficacy of our proposed spherical fusion based encoder $\Phi$, we compare with three alternative encoders: (1) a baseline of \textbf{Densefusion} \cite{Densefusion}, a pose encoder that fuses the learned RGB features from CNNs and point features from PointNet \cite{PointNet} in a point-wise manner; (2) \textbf{SCNN-EarlyFusion}, which takes as input the concatenation of $\mathcal{S}^{\mathcal{X}}$ and $\mathcal{S}^{\mathcal{P}}$ and feeds it into a multi-scale spherical CNN, followed by an MLP; (3) \textbf{SCNN-LateFusion}, which first feeds $\mathcal{S}^{\mathcal{X}}$ and $\mathcal{S}^{\mathcal{P}}$ into two separate multi-scale spherical CNNs and applies an MLP to the concatenation of the two output features. The used multi-scale spherical CNN is constructed by $8$ spherical convolution layers, with aggregation of multi-scale spherical features similar to $\Phi$.
We conduct ablation experiments by replacing $\Phi$ with the above encoders, while keeping $\Psi_{exp}$ and $\Psi_{im}$ as remained.
Results (without the refined learning of pose consistency) in Table \ref{Table:Ablation} show that the three alternative encoders perform worse than our proposed $\Phi$ with spherical fusion. Compared with the densefusion baseline, those based on spherical convolutions enjoy the property of rotation equivariance on $SO(3)$, and thus achieve higher mAPs. With spherical fusion, our proposed pose encoder $\Phi$ enables information communication progressively along the hierarchy, outperforming either SCNN-EarlyFusion with feature fusion at the very beginning or SCNN-LateFusion at the end.

\begin{figure}[t]
    \begin{center}
    \vspace{-0.1cm}
       \includegraphics[width=0.6\linewidth]{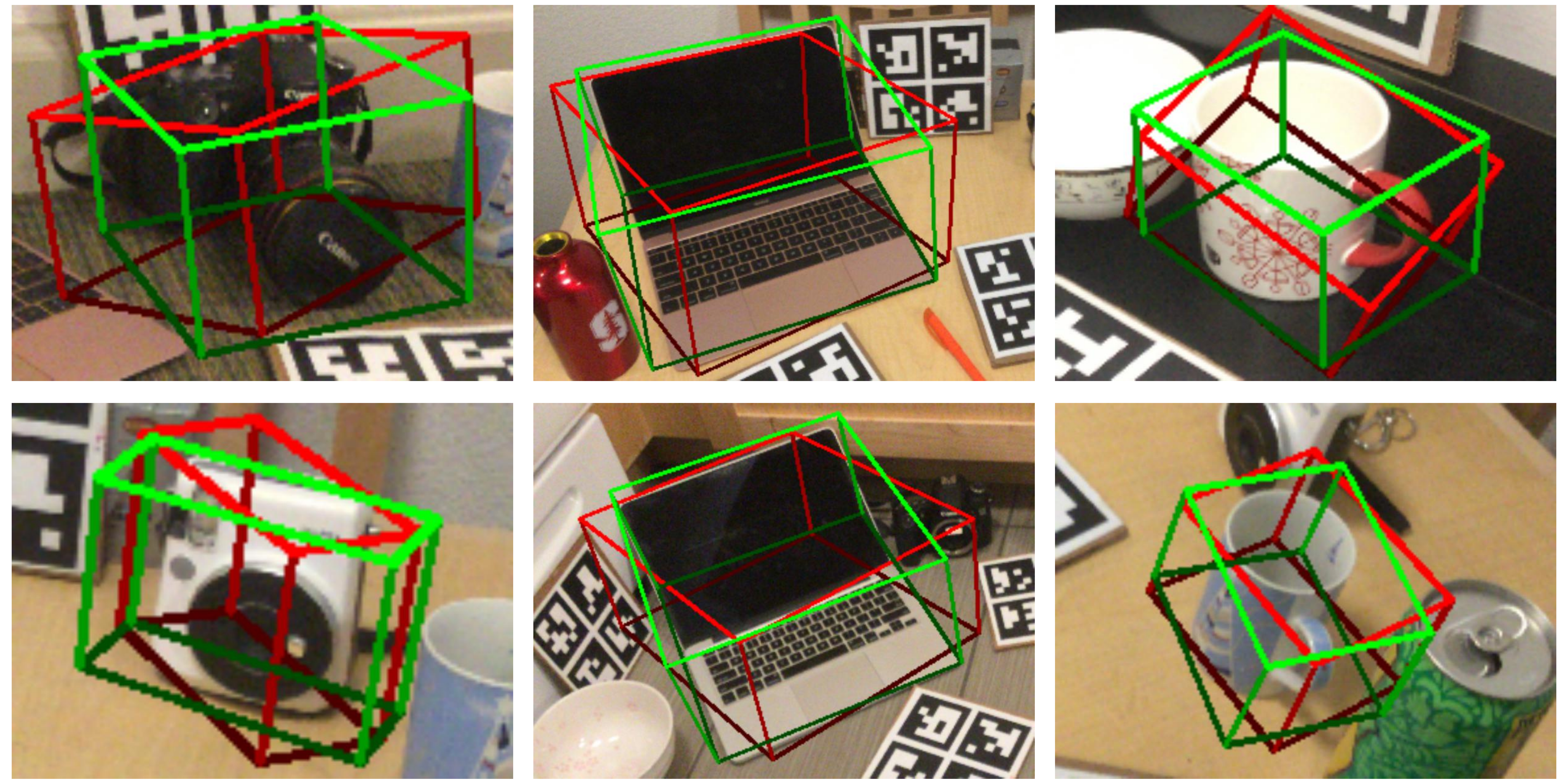}
    \end{center}
    \vspace{-0.2cm}
       \caption{Qualitative results of DualPoseNet without (red) and with (green) the refined learning of pose consistency on REAL275. }
    \label{fig:Refinement_vis}
    \vspace{-0.4cm}
\end{figure}

\begin{figure}[t]
   \begin{center}
   \includegraphics[width=0.55\linewidth]{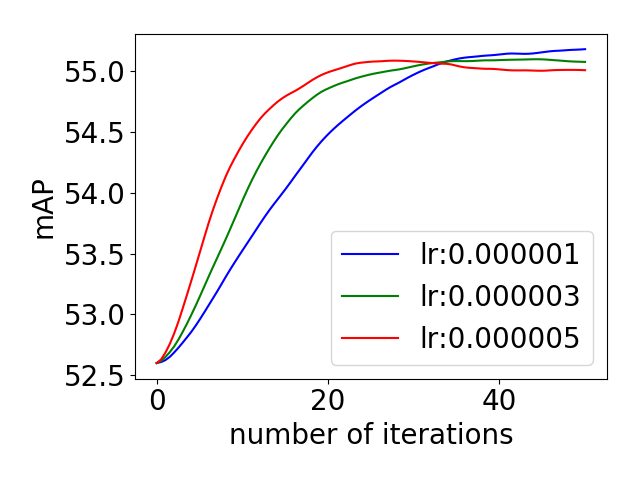}
   \end{center}
   \vspace{-0.6cm}
      \caption{ Plottings of prediction accuracy (mAP of (IoU$_{50}$, $10\degree$, $20\%$)) versus the number of iterations when using different learning
   rates to fine-tune the loss (\ref{EqnLossRefine}) for the refined learning of pose consistency. Experiments are conducted on REAL275 \cite{NOCS}.}
   \label{fig:cues}
   \vspace{-0.4cm}
\end{figure}

\begin{table*}
   \begin{center}
       \vspace{-0.2cm}
       \resizebox{\textwidth}{!}{
       \begin{tabular}{c|c|cccccc|cccccc}
           \hline
           \multirow{3}{*}{Dataset} & \multirow{3}{*}{Method} & \multicolumn{12}{c}{mAP} \\
           \cline{3-14}
           & & IoU$_{75}$ & IoU$_{75}$ & IoU$_{75}$ & IoU$_{50}$ & IoU$_{50}$ & IoU$_{50}$ & \multirow{2}{*}{IoU$_{50}$} & \multirow{2}{*}{IoU$_{75}$} & 5\degree & 5\degree & 10\degree & 10\degree\\
           & & 5\degree, $5\%$ & 10\degree, $5\%$ & 5\degree, $10\%$ & 5\degree, $20\%$ & 10\degree, $10\%$ & 10\degree, $20\%$ & & & 2cm & 5cm & 2cm & 5cm \\

           \hline
           \hline
           \multirow{3}{*}{CAMERA25} & NOCS \cite{NOCS}      & $22.6$          & $29.5$          & $31.5$          & $34.5$          & $54.5$          & $56.8$        & $83.9$          & $69.5$          & $32.3$          & $40.9$          & $48.2$          & $64.6$\\
                                     & SPD \cite{ShapePrior} & $47.5$      & $61.5$          & $52.2$          & $56.6$          & $75.3$          & $78.5$    & $\mathbf{93.2}$ & $83.1$          & $54.3$          & $59.0$          & $73.3$          & $81.5$\\
           \cline{2-14}
              & DualPoseNet                           & $\mathbf{56.2}$ & $\mathbf{65.1}$ & $\mathbf{65.1}$ & $\mathbf{68.0}$ & $\mathbf{78.6}$ & $\mathbf{81.5}$& $92.4$          & $\mathbf{86.4}$ & $\mathbf{64.7}$ & $\mathbf{70.7}$ & $\mathbf{77.2}$ & $\mathbf{84.7}$\\
           \hline
           \hline
           \multirow{4}{*}{REAL275} & NOCS \cite{NOCS}      & $2.4$         & $3.5$          & $7.1$           & $9.3$          & $19.7$          & $22.3$         & $78.0$         & $30.1$          & $7.2$           & $10.0$          & $13.8$          & $25.2$         \\
           & SPD \cite{ShapePrior} & $8.6$         & $17.2$          & $15.0$          & $17.4$          & $38.5$          & $42.5$         & $77.3$         & $53.2$          & $19.3$          & $21.4$          & $43.2$          & $54.1$         \\
           & CASS \cite{CASS} & $-$         & $-$          & $-$          & $-$          & $-$          & $-$         & $77.7$         & $-$          & $-$          & $23.5$          & $-$          & $58.0$         \\
           \cline{2-14}
              & DualPoseNet                          & $\mathbf{11.2}$& $\mathbf{17.2}$ & $\mathbf{24.8}$ & $\mathbf{29.8}$ & $\mathbf{44.5}$ & $\mathbf{55.0}$ & $\mathbf{79.8}$& $\mathbf{62.2}$ & $\mathbf{29.3}$ & $\mathbf{35.9}$ & $\mathbf{50.0}$ & $\mathbf{66.8}$ \\
           \hline
       \end{tabular}}
   \end{center}
   \vspace{-0.1cm}
   \caption{Quantitative comparisons of different methods on CAMERA25 and REAL275. Evaluations are based on both our proposed metrics (left) and the metrics (right) proposed in \cite{NOCS}.}\label{Table:SOTA}
\end{table*}

\begin{figure*}[t]
   \begin{center}
   \vspace{-0.2cm}
      \includegraphics[width=1.0\linewidth]{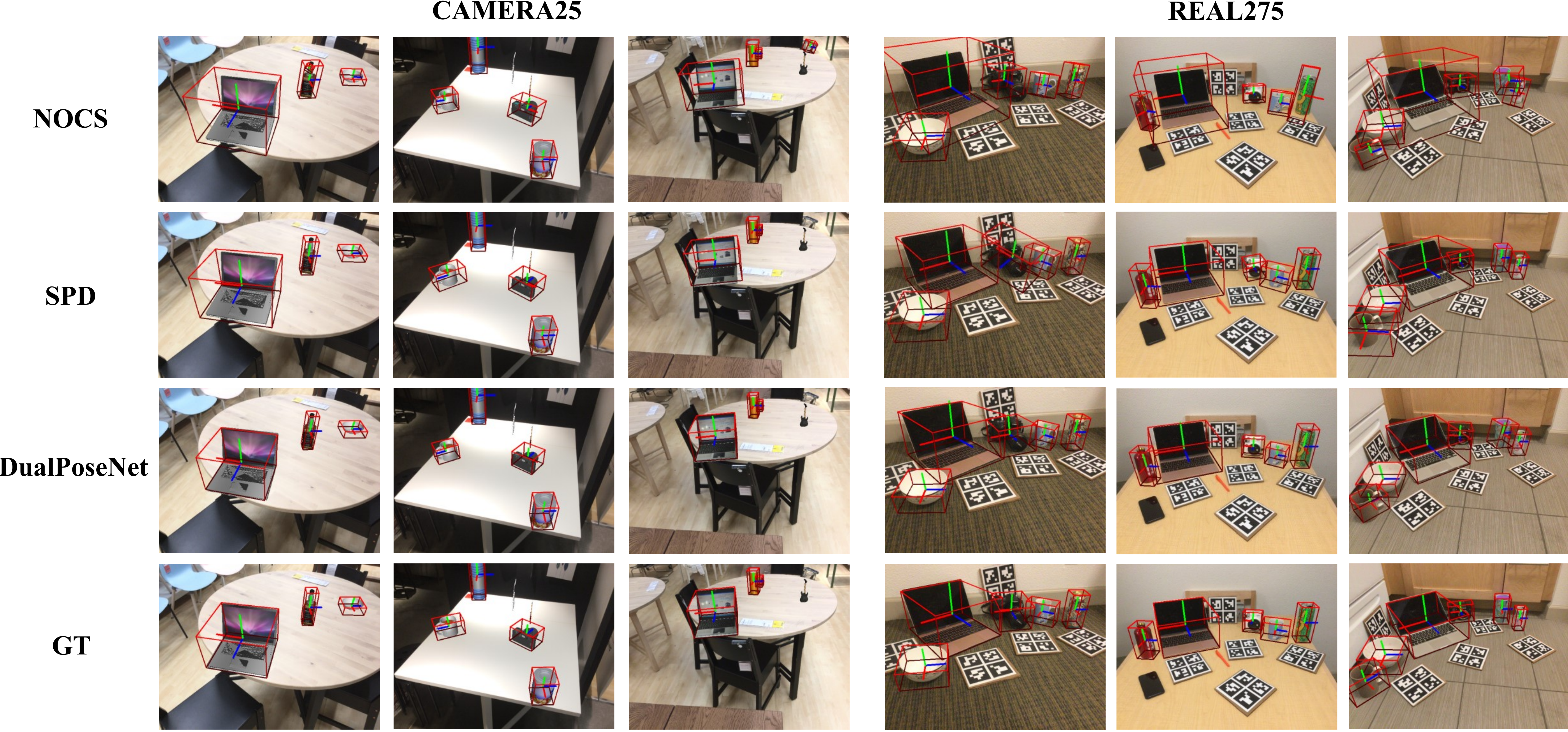}
   \end{center}
   \vspace{-0.3cm}
      \caption{Qualitative results of different methods on CAMERA25 and REAL275 \cite{NOCS}. }
   \label{fig:SOTA}
   \vspace{-0.2cm}
\end{figure*}

We finally investigate the benefit from the proposed refined learning of pose consistency. Results in Table \ref{Table:Ablation} show that with the refined learning, pose precisions improve stably across the full spectrum of evaluation metrics, and the improvements increase when coarser metrics are used; this suggests that the refinement indeed attracts the learning towards more accurate regions in the solution space of pose prediction. 
Examples in Fig. \ref{fig:Refinement_vis} give corroborative evidence of the efficacy of the refined learning.
In fact, the refinement process is a trade-off between the improved precision and refining efficiency. As mentioned in Section \ref{SecDualPoseNetRefine}, the refining efficiency depends on the learning rate and number of iterations for fine-tuning the encoder with the objective (\ref{EqnLossRefine}). In Fig. \ref{fig:cues}, we plot curves of mAP of (IoU$_{50}$, $10\degree$, $20\%$) against the number of iterations when using different learning rates.
It shows faster convergence when using larger learning rates, which, however, may end with less mature final results, even with overfitting.
In practice, one may set a proper tolerance $\epsilon$ for a balanced efficiency and accuracy. We set the learning rate as $1\times10^{-6}$ and $\epsilon = 5\times10^{-5}$ for results of DualPoseNet reported in the present paper. Under this setting, it costs negligible $0.2$ seconds per instance on a server with Intel E5-2683 CPU and GTX 1080ti GPU.

\vspace{-0.2cm}
\subsubsection{Comparisons with Existing Methods} \label{Sec:SOTA}

We compare our proposed DualPoseNet with the existing methods, including NOCS \cite{NOCS}, SPD \cite{ShapePrior}, and CASS \cite{CASS},  on the CAMERA25 and REAL275 \cite{NOCS} datasets. Note that NOCS and SPD are designed to first predict canonical versions of the observed point clouds, and the poses are obtained from post-alignment by solving a Umeyama algorithm \cite{umeyama1991least}. 
Quantitative results in Table \ref{Table:SOTA} show the superiority of our proposed DualPoseNet on both datasets, especially for the metrics of high precisions. 
For completeness, we also present in Table \ref{Table:SOTA} the comparative results under the original evaluation metrics proposed in \cite{NOCS}; our results are better than existing ones at all but one rather coarse metric of IoU$_{50}$, which is in fact a metric less sensitive to object pose. 
Qualitative results of different methods are shown in Fig. \ref{fig:SOTA}. Comparative advantages of our method over the existing ones are consistent with those observed in Table \ref{Table:SOTA}. For example, the bounding boxes of laptops in the figure generated by NOCS and SPD are obviously bigger than the exact extensions of laptops, while our method predicts more compact bounding boxes with precise poses and sizes. More comparative results are shown in the supplementary material.

\subsection{Instance-level 6D Pose Estimation} \label{subsec:exp-instance-6d}

\begin{table}[t]
       \centering
      \resizebox{0.48\textwidth}{!}{
       \begin{tabular}{cccc|cc}
       \hline
       Encoder & $\Psi_{im}$ & Refining & Iterative  & YCB-Video & LineMOD \\
       \hline
      Densefusion \cite{Densefusion} & $\times$ & $\times$ & $\times$ &  $88.2$ & $78.7$\\
       \hline
        $\Phi$ & $\times$ & $\times$ & $\times$ &  $90.5$ & $88.6$\\
        $\Phi$ & \checkmark & $\times$ &$\times$ & $91.2$ & $92.7$\\
        $\Phi$ & \checkmark & \checkmark  & $\times$& $93.3$  &  $94.6$\\
        $\Phi$ & \checkmark & $\times$ & \checkmark  & $95.0$ &  $96.3$\\
        $\Phi$ & \checkmark & \checkmark  & \checkmark  & $\mathbf{96.5}$ & $\mathbf{98.2}$\\
       \hline
      \end{tabular}}
      \vspace{0.05cm}
      \caption{Ablation studies on variants of DualPoseNet on YCB-Video \cite{YCB} and LineMOD \cite{LineMod} datasets for instance-level 6D pose estimation. The evaluation metrics are mean ADD-S AUC and mean ADD(S) AUC, respectively.}
       \label{tab:exp_instance_ablation}
       \vspace{-0.3cm}
\end{table}

\begin{table}[t]
       \centering
      \resizebox{0.45\textwidth}{!}{
       \begin{tabular}{c|cc}
       \hline
       Method  & YCB-Video & LineMOD \\
       \hline
       Pointfusion \cite{Pointfusion} & $83.9$ & $73.7$ \\
       PoseCNN + ICP \cite{xiang2017posecnn} & $93.0$ & $-$ \\
       Densefusion (Per-pixel) \cite{Densefusion}  & $91.2$ & $86.2$ \\
       Densefusion (Iterative) \cite{Densefusion}  & $93.1$ & $94.3$ \\
       $\mathcal{W}$-PoseNet \cite{xu2019w} & $93.0$ & $97.2$ \\
       $\mathcal{W}$-PoseNet (Iterative)\cite{xu2019w} & $94.0$ & $98.1$ \\
       \hline
       DualPoseNet  & $93.3$ & $94.6$ \\
       DualPoseNet (Iterative)  & $\mathbf{96.5}$ & $\mathbf{98.2}$ \\
       \hline
      \end{tabular}}
      \vspace{0.2cm}
      \caption{Quantitative comparisons of different methods on YCB-Video \cite{YCB} and LineMOD \cite{LineMod} datasets for instance-level 6D pose estimation. The evaluation metrics are mean ADD-S AUC and mean ADD(S) AUC, respectively.
      }
      \label{tab:exp_instance_sota}
      \vspace{-0.5cm}
\end{table}

We apply DualPoseNet to YCB-Video \cite{YCB} and LineMOD \cite{LineMod} datasets for the instance-level task. Results in Table \ref{tab:exp_instance_ablation} confirm the efficacy of our individual components (the encoder $\Phi$, the implicit decoder ${\Psi}_{im}$, and the refined learning of pose consistency); following \cite{Densefusion, xu2019w}, we also augment our DualPoseNet with a 2nd-stage module for iterative refinement of residual pose, denoted as DualPoseNet(Iterative), to further improve the performance. As shown in Table \ref{tab:exp_instance_sota}, DualPoseNet(Iterative) achieves comparable results against other methods, showing its potential for use in instance-level tasks. More quantitative and qualitative results are shown in the supplementary material.

\section{Acknowledgement}
This work was partially supported by the Guangdong R$\&$D key project of China (No.: 2019B010155001), the National Natural Science Foundation of China (No.: 61771201), and the Program for Guangdong Introducing Innovative and Entrepreneurial Teams (No.: 2017ZT07X183).

{\small
\bibliographystyle{ieee_fullname}
\bibliography{egbib}

\begin{thebibliography}{10}\itemsep=-1pt

\bibitem{matterport_maskrcnn_2017}
Waleed Abdulla.
\newblock Mask r-cnn for object detection and instance segmentation on keras
  and tensorflow.
\newblock \url{https://github.com/matterport/Mask_RCNN}, 2017.

\bibitem{ICP}
Paul~J Besl and Neil~D McKay.
\newblock Method for registration of 3-d shapes.
\newblock In {\em Sensor fusion IV: control paradigms and data structures},
  volume 1611, pages 586--606. International Society for Optics and Photonics,
  1992.

\bibitem{YCB}
Berk Calli, Arjun Singh, Aaron Walsman, Siddhartha Srinivasa, Pieter Abbeel,
  and Aaron~M Dollar.
\newblock The ycb object and model set: Towards common benchmarks for
  manipulation research.
\newblock In {\em 2015 international conference on advanced robotics (ICAR)},
  pages 510--517. IEEE, 2015.

\bibitem{ShapeNet}
Angel~X Chang, Thomas Funkhouser, Leonidas Guibas, Pat Hanrahan, Qixing Huang,
  Zimo Li, Silvio Savarese, Manolis Savva, Shuran Song, Hao Su, et~al.
\newblock Shapenet: An information-rich 3d model repository.
\newblock {\em arXiv preprint arXiv:1512.03012}, 2015.

\bibitem{CASS}
Dengsheng Chen, Jun Li, and Kai Xu.
\newblock Learning canonical shape space for category-level 6d object pose and
  size estimation.
\newblock In {\em Conference on Computer Vision and Pattern Recognition
  (CVPR)}, 2020.

\bibitem{chen2020category}
Xu Chen, Zijian Dong, Jie Song, Andreas Geiger, and Otmar Hilliges.
\newblock Category level object pose estimation via neural
  analysis-by-synthesis.
\newblock In {\em European Conference on Computer Vision}, pages 139--156.
  Springer, 2020.

\bibitem{SCNNICLR}
Taco~S. Cohen, Mario Geiger, Jonas Koehler, and Max Welling.
\newblock Spherical cnns.
\newblock In {\em ICLR}, 2018.

\bibitem{drost2010model}
Bertram Drost, Markus Ulrich, Nassir Navab, and Slobodan Ilic.
\newblock Model globally, match locally: Efficient and robust 3d object
  recognition.
\newblock In {\em CVPR}, 2010.

\bibitem{SCNN}
Carlos Esteves, Christine Allen-Blanchette, Ameesh Makadia, and Kostas
  Daniilidis.
\newblock Learning so (3) equivariant representations with spherical cnns.
\newblock In {\em Proceedings of the European Conference on Computer Vision
  (ECCV)}, pages 52--68, 2018.

\bibitem{KITTI}
Andreas Geiger, Philip Lenz, and Raquel Urtasun.
\newblock Are we ready for autonomous driving? the kitti vision benchmark
  suite.
\newblock In {\em Conference on Computer Vision and Pattern Recognition
  (CVPR)}, 2012.

\bibitem{MaskRCNN}
Kaiming He, Georgia Gkioxari, Piotr Doll{\'a}r, and Ross Girshick.
\newblock Mask r-cnn.
\newblock In {\em Proceedings of the IEEE international conference on computer
  vision}, pages 2961--2969, 2017.

\bibitem{PoseTemplateMatchingPAMI}
Stefan Hinterstoisser, C{\'e}dric Cagniart, Slobodan Ilic, Peter Sturm, Nassir
  Navab, Pascal Fua, and Vincent Lepetit.
\newblock {Gradient Response Maps for Real-Time Detection of Texture-Less
  Objects}.
\newblock {\em {IEEE Transactions on Pattern Analysis and Machine
  Intelligence}}, 34(5):876--888, 2012.

\bibitem{LineMod}
Stefan Hinterstoisser, Stefan Holzer, Cedric Cagniart, Slobodan Ilic, Kurt
  Konolige, Nassir Navab, and Vincent Lepetit.
\newblock Multimodal templates for real-time detection of texture-less objects
  in heavily cluttered scenes.
\newblock In {\em 2011 international conference on computer vision}, pages
  858--865. IEEE, 2011.

\bibitem{hinterstoisser2016going}
Stefan Hinterstoisser, Vincent Lepetit, Naresh Rajkumar, and Kurt Konolige.
\newblock Going further with point pair features.
\newblock In {\em ECCV}, 2016.

\bibitem{ShapeCollectionAlignment}
Qi-Xing Huang, Hao Su, and Leonidas Guibas.
\newblock Fine-grained semi-supervised labeling of large shape collections.
\newblock {\em ACM Transactions on Graphics (TOG)}, 2013.

\bibitem{kehl2017ssd}
Wadim Kehl, Fabian Manhardt, Federico Tombari, Slobodan Ilic, and Nassir Navab.
\newblock Ssd-6d: Making rgb-based 3d detection and 6d pose estimation great
  again.
\newblock In {\em ICCV}, 2017.

\bibitem{li2018unified}
Chi Li, Jin Bai, and Gregory~D Hager.
\newblock A unified framework for multi-view multi-class object pose
  estimation.
\newblock In {\em Proceedings of the European Conference on Computer Vision
  (ECCV)}, pages 254--269, 2018.

\bibitem{DeepIm}
Yi Li, Gu Wang, Xiangyang Ji, Yu Xiang, and Dieter Fox.
\newblock Deepim: Deep iterative matching for 6d pose estimation.
\newblock In {\em Proceedings of the European Conference on Computer Vision
  (ECCV)}, pages 683--698, 2018.

\bibitem{manhardt2020cps++}
Fabian Manhardt, Gu Wang, Benjamin Busam, Manuel Nickel, Sven Meier, Luca
  Minciullo, Xiangyang Ji, and Nassir Navab.
\newblock Cps++: Improving class-level 6d pose and shape estimation from
  monocular images with self-supervised learning.
\newblock {\em arXiv preprint arXiv:2003.05848}, 2020.

\bibitem{peng2019pvnet}
Sida Peng, Yuan Liu, Qixing Huang, Xiaowei Zhou, and Hujun Bao.
\newblock Pvnet: Pixel-wise voting network for 6dof pose estimation.
\newblock In {\em CVPR}, 2019.

\bibitem{VoteNet}
Charles~R Qi, Or Litany, Kaiming He, and Leonidas~J Guibas.
\newblock Deep hough voting for 3d object detection in point clouds.
\newblock In {\em Proceedings of the IEEE International Conference on Computer
  Vision}, pages 9277--9286, 2019.

\bibitem{FPointNet}
Charles~R Qi, Wei Liu, Chenxia Wu, Hao Su, and Leonidas~J Guibas.
\newblock Frustum pointnets for 3d object detection from rgb-d data.
\newblock In {\em Proceedings of the IEEE Conference on Computer Vision and
  Pattern Recognition}, pages 918--927, 2018.

\bibitem{PointNet}
Charles~R Qi, Hao Su, Kaichun Mo, and Leonidas~J Guibas.
\newblock Pointnet: Deep learning on point sets for 3d classification and
  segmentation.
\newblock In {\em Proceedings of the IEEE conference on computer vision and
  pattern recognition}, pages 652--660, 2017.

\bibitem{PointRCNN}
Shaoshuai Shi, Xiaogang Wang, and Hongsheng Li.
\newblock Pointrcnn: 3d object proposal generation and detection from point
  cloud.
\newblock In {\em Proceedings of the IEEE Conference on Computer Vision and
  Pattern Recognition}, pages 770--779, 2019.

\bibitem{SUNRGBD}
Shuran Song, Samuel~P Lichtenberg, and Jianxiong Xiao.
\newblock Sun rgb-d: A rgb-d scene understanding benchmark suite.
\newblock In {\em Proceedings of the IEEE conference on computer vision and
  pattern recognition}, pages 567--576, 2015.

\bibitem{sundermeyer2018implicit}
Martin Sundermeyer, Zoltan-Csaba Marton, Maximilian Durner, Manuel Brucker, and
  Rudolph Triebel.
\newblock Implicit 3d orientation learning for 6d object detection from rgb
  images.
\newblock In {\em ECCV}, 2018.

\bibitem{ShapePrior}
Meng Tian, Marcelo~H Ang~Jr, and Gim~Hee Lee.
\newblock Shape prior deformation for categorical 6d object pose and size
  estimation.
\newblock In {\em Proceedings of the European Conference on Computer Vision
  (ECCV)}, August 2020.

\bibitem{umeyama1991least}
Shinji Umeyama.
\newblock Least-squares estimation of transformation parameters between two
  point patterns.
\newblock {\em IEEE Transactions on Pattern Analysis \& Machine Intelligence},
  (4):376--380, 1991.

\bibitem{Densefusion}
Chen Wang, Danfei Xu, Yuke Zhu, Roberto Mart{\'\i}n-Mart{\'\i}n, Cewu Lu, Li
  Fei-Fei, and Silvio Savarese.
\newblock Densefusion: 6d object pose estimation by iterative dense fusion.
\newblock In {\em Proceedings of the IEEE Conference on Computer Vision and
  Pattern Recognition}, pages 3343--3352, 2019.

\bibitem{NOCS}
He Wang, Srinath Sridhar, Jingwei Huang, Julien Valentin, Shuran Song, and
  Leonidas~J Guibas.
\newblock Normalized object coordinate space for category-level 6d object pose
  and size estimation.
\newblock In {\em Proceedings of the IEEE Conference on Computer Vision and
  Pattern Recognition}, pages 2642--2651, 2019.

\bibitem{FConvNet}
Zhixin Wang and Kui Jia.
\newblock Frustum convnet: Sliding frustums to aggregate local point-wise
  features for amodal 3d object detection.
\newblock In {\em The IEEE/RSJ International Conference on Intelligent Robots
  and Systems (IROS)}, 2019.

\bibitem{xiang2017posecnn}
Yu Xiang, Tanner Schmidt, Venkatraman Narayanan, and Dieter Fox.
\newblock Posecnn: A convolutional neural network for 6d object pose estimation
  in cluttered scenes.
\newblock 2018.

\bibitem{Pointfusion}
Danfei Xu, Dragomir Anguelov, and Ashesh Jain.
\newblock Pointfusion: Deep sensor fusion for 3d bounding box estimation.
\newblock In {\em Proceedings of the IEEE Conference on Computer Vision and
  Pattern Recognition}, pages 244--253, 2018.

\bibitem{xu2019w}
Zelin Xu, Ke Chen, and Kui Jia.
\newblock W-posenet: Dense correspondence regularized pixel pair pose
  regression.
\newblock {\em arXiv preprint arXiv:1912.11888}, 2019.

\bibitem{STD}
Zetong Yang, Yanan Sun, Shu Liu, Xiaoyong Shen, and Jiaya Jia.
\newblock Std: Sparse-to-dense 3d object detector for point cloud.
\newblock In {\em Proceedings of the IEEE International Conference on Computer
  Vision}, pages 1951--1960, 2019.

\bibitem{VoxelNet}
Yin Zhou and Oncel Tuzel.
\newblock Voxelnet: End-to-end learning for point cloud based 3d object
  detection.
\newblock In {\em Proceedings of the IEEE Conference on Computer Vision and
  Pattern Recognition}, pages 4490--4499, 2018.

\end{thebibliography}
}

\end{document}